\documentclass[letterpaper, 10pt, conference]{ieeeconf}      

\IEEEoverridecommandlockouts                         

\usepackage{latexsym}
\usepackage{amssymb}
\usepackage{amsmath}
\usepackage{graphicx}
\usepackage{color}
\usepackage{listings}
\usepackage{hyperref}
\usepackage{url}
\usepackage{cleveref}
\usepackage{algorithm}
\usepackage{algpseudocode}
\usepackage{subcaption}
\usepackage{tabularx}
\usepackage{multirow}
\usepackage[colorinlistoftodos]{todonotes}

\newtheorem{definition}{Definition}

\usepackage{xcolor}
\usepackage[utf8]{inputenc}

\lstdefinelanguage{PDDL}
{
  sensitive=false,    
  morecomment=[l]{;}, 
  alsoletter={:,-},   
  morekeywords={
    define,domain,problem,not,and,or,when,forall,exists,either,
    :domain,:requirements,:types,:objects,:constants,
    :predicates,:action,:parameters,:precondition,:effect,
    :fluents,:primary-effect,:side-effect,:init,:goal,
    :strips,:adl,:equality,:typing,:conditional-effects,
    :negative-preconditions,:disjunctive-preconditions,
    :existential-preconditions,:universal-preconditions,:quantified-preconditions,
    :functions,assign,increase,decrease,scale-up,scale-down,
    :metric,minimize,maximize,
    :durative-actions,:duration-inequalities,:continuous-effects,
    :durative-action,:duration,:condition
  }
}

\crefname{lstlisting}{listing}{listings}
\Crefname{lstlisting}{Listing}{Listings}

\title{\LARGE \bf
GraSP-VLA: Graph-based Symbolic Action Representation for Long-Horizon Planning with VLA Policies
}

\author{
Maëlic Neau$^{1}$, Zoe Falomir$^{1}$, Paulo E. Santos$^{2}$,  Anne-Gwenn Bosser$^{3}$ and Cédric Buche$^{4,5}$
\thanks{$^{1}$Computing Science Department, Umeå University, Umeå, Sweden}
\thanks{$^{2}$PrioriAnalytica, Adelaide, Australia}%
\thanks{$^{3}$Bretagne INP - ENIB, Plouzané, France}%
\thanks{$^{4}$IMT Atlantique, Plouzané, France}%
\thanks{$^{5}$CNRS IRL 2010 CROSSING, Adelaide, Australia}%
}

\begin{document}

\maketitle
\thispagestyle{empty}
\pagestyle{empty}

\begin{abstract}

Deploying autonomous robots that can learn new skills from demonstrations is an important challenge of modern robotics. Existing solutions often apply end-to-end imitation learning with Vision-Language Action (VLA) models or symbolic approaches with Action Model Learning (AML). On the one hand, current VLA models are limited by the lack of high-level symbolic planning, which hinders their abilities in long-horizon tasks. On the other hand, symbolic approaches in AML lack generalization and scalability perspectives.
In this paper we present a new neuro-symbolic approach, GraSP-VLA, a framework that uses a Continuous Scene Graph representation to generate a symbolic representation of human demonstrations. This representation is used to generate new planning domains during inference and serves as an orchestrator for low-level VLA policies, scaling up the number of actions that can be reproduced in a row.
Our results show that GraSP-VLA is effective for modeling symbolic representations on the task of automatic planning domain generation from observations. In addition, results on real-world experiments show the potential of our Continuous Scene Graph representation to orchestrate low-level VLA policies in long-horizon tasks.
\end{abstract}

\section{Introduction}


Inferring the preconditions and outcomes of actions from observations is a long-lasting challenge in robotics. Action Model Learning (AML) \cite{aroraReviewLearningPlanning2018b} addresses this challenge by modeling symbolic representations of atomic actions from visual observations of human demonstrations. These representations can then be used to compose long-horizon tasks using pre-trained low-level behaviors.
In AML, we define an action as a set of initial states (i.e. \textit{preconditions}), transitions, and final states (i.e. \textit{effects}), defined by a collection of known predicates. To account for direct changes in the environment, the sets of initial and final states should be disjoint, as a result of the action as a direct modification to the state of the world.

A major bottleneck of AML approaches, and by extension of symbolic representations in robotics, is the cost of human labor involved in designing the representations \cite{zanchettinEndtoendActionModel2025}. In fact, AML relies on domain knowledge from human experts to collect predicates, hindering generalization and scalability perspectives.
Recently, Vision-Language Models (VLMs) have emerged as low-cost experts to gather domain knowledge \cite{zhangDKPROMPTDomainKnowledge2024}. However, the use of VLMs in robotics involves new challenges, such as inference cost in low-resource and time-constrained applications. Hallucinations and a lack of symbol grounding are also issues of VLMs that require consideration for their applications to robotics \cite{faveroMultiModalHallucinationControl2024}. 
In the past few years, we have also seen the emergence of Scene Graph Generation (SGG) \cite{xuSceneGraphGeneration2017}, a task that aims to model symbolic representations of raw images as graph structures. At first glance, the similarities between SGG and AML are scarce. However, after a closer look, opportunities can emerge. First, scene graph representations are composed of (1) collections of objects grounded to the scene and (2) sets of predicates that form visual relationships (i.e. also known as \textit{triplets}). Second, SGG methods are based on supervised learning, which leverages the use of large-scale, diverse databases of images \cite{krishnaVisualGenomeConnecting2017a}.
In summary, SGG approaches can describe the interplay of humans, physical objects, and background elements with fine-grained details in a wide variety of contexts. In contrast to VLMs, SGG models are light-weight, do not hallucinate, and directly ground the symbolic representation to the visual content, which makes them good candidates for the autonomous extraction of action descriptions in AML.

Although applying SGG methods to AML could lead to a significant leap forward, related literature is extremely scarce \cite{amodeoOGSGGOntologyGuidedScene2022}. In this work, we aim to solve this gap by proposing the first implementation of SGG in the context of Action Model Learning. As an overview of the method proposed in this work, SGG approaches are used as a backbone to extract key information from the visual content as a graph structure. This representation is then aggregated over time in what we introduce as a \textit{Multi-Layer Continuous Scene Graph (ML-CSG)}, which serves as the internal memory of the autonomous agent. This representation is used to automatically extract domain knowledge in the form of Planning Domain Description Language (PDDL) actions. PDDL actions can be used in combination with traditional planning solvers in goal-based Imitation Learning. Thus, in this work, we also propose to show the advantage of our approach in the context of Open-Ended Imitation Learning. The main difference is that, in Open-Ended Imitation Learning, the goal of the task is not known. As a result, atomic actions can be composed online, allowing for the immediate reproduction of the task, without waiting for the end of the demonstration.

We provide a set of experiments using our approach for Open-Ended Imitation Learning with a collection of low-level Vision-Language Action model (VLA) policies \cite{kim2025openvla}, previously pre-trained on a small set of simple behaviors.
VLA models are a recent yet promising paradigm for autonomous manipulation, where the focus is on the ease of deployment and generalization to new environments \cite{shukor2025smolvla}. A major concern with current VLA models is the significant drop in performance in long-horizon tasks \cite{merler2025viplan}. Instead of training a single policy on a long and complex task, we propose to train a collection of individual policies on a set of simple, atomic behaviors which can be composed at inference time with our \textit{ML-CSG} to reproduce the entire task. 

This paper combines SGG, Multi-Layer Continuous Scene Graph, domain knowledge extraction, and scheduling of the VLA policies as the Graph-based Symbolic Planning for Vision-Language Action models (GraSP-VLA) architecture.
Our contributions can be summarized as follows:
\begin{enumerate}
       \item A new approach to SGG with the extension of the Scene Graph representation to a Multi-Layer Continuous Scene Graph;
       \item A new algorithm for the automatic generation of planning domains from the Continuous Scene Graph;
       \item A Continuous Scene Graph-based orchestrator to decompose tasks into sequences of atomic behaviors; and
        \item A client-server execution using a bank of pre-trained Vision-Language Action model policies.
\end{enumerate}

\section{Related Work}

\textbf{Imitation learning} \cite{fangSurveyImitationLearning2019}, also defined as the transfer of human skills to robots, has been a growing research area in the past few years. While traditional methods may include Reinforcement Learning (RL) to learn trajectories \cite{piotBridgingGapImitation2017}, symbolic approaches have been overlooked \cite{ramirezTransferringSkillsHumanoid2017}. Symbolic imitation learning aims to learn new tasks as high-level, structured representations (symbols) of states, actions, and goals, rather than raw sensor data. This includes, for instance, symbol-grounding networks for policy generation \cite{huangContinuousRelaxationSymbolic2019a}. States and actions can also be learned through the automatic generation of planning domains \cite{diehlAutomatedGenerationRobotic2021}, a process we refer to as Action Model Learning (AML) \cite{aroraReviewLearningPlanning2018b}.
In a recent work, Zanchettin \cite{zanchettinSymbolicRepresentationWhat2023} explores the usage of semantic graphs for planning domain generation in AML. The introduced semantic graphs are limited to relative spatial relationships with known positions extracted by calibration tags. In addition, Zanchettin \cite{zanchettinSymbolicRepresentationWhat2023} defines predicates in advance, which is not generalizable. In this paper, we aim to show that the autonomous generation of a planning domain can be generalized to the real world through the use of a dedicated Scene Graph representation. 
Our approach GraSP-VLA bridges the perception-to-reasoning gap by generating a comprehensive planning domain from human demonstrations without any explicit priors.

\textbf{Scene Graph Generation (SGG)} aims to model visual relationships extracted from images into grounded graph structures \cite{xuSceneGraphGeneration2017}. A SGG model first detects relevant object regions in the form of bounding boxes (i.e. object detection) and then performs relation prediction on the set of detected object pairs. This way, relations can be grounded to the scene via the corresponding $\langle subject, object \rangle$ regions.
RCNN-style networks were among the earliest approaches applied within SGG research \cite{zellersNeuralMotifsScene2018a} to learn visual representations of both objects and relationships in a two-stage pipeline. Recently, one-stage approaches have been proposed, leveraging the DETR architecture \cite{imEGTRExtractingGraph2024} to learn both relation and object representations jointly. One of the main concerns of both two-stage and one-stage approaches is a lack of efficient real-time implementation, which is a significant constraint for real-world robotic implementation. A recent work [anonymized] proposes a real-time and lightweight SGG model, REACT, based on the latest YOLO architectures \cite{Jocher_Ultralytics_YOLO_2023}, which presents a good trade-off between latency, object detection accuracy, and relation prediction accuracy for real-world robotics applications. However, SGG models can only generate scene graphs from still images. In this paper, we introduce the concept of a Continuous Scene Graph where nodes and edges are persistent through time. This representation then serves as the internal memory of a robotic agent.

\textbf{Vision-Language Action models (VLAs)} \cite{kim2025openvla} have recently gained interest in the robotics community. VLAs can learn manipulation policies through vision and language supervision from demonstrations \cite{shukor2025smolvla}. However, the main concern lies in their ability to maintain high performance in long-horizon tasks \cite{zhang2024vlabench}. To approach this challenge, we propose a new paradigm for learning by demonstration: Continuous Scene Graph as Orchestrator for VLA Policies. Instead of training a single VLA policy on a complex task, we can decouple the task into smaller chunks (low-level unitary skills) and train different policies for each skill. Then, to reproduce the full task, we can leverage our Continuous Scene Graph to extract the correct scheduling.

\section{GraSP-VLA Architecture}

\begin{figure*}
    \centering
    \includegraphics[width=\textwidth]{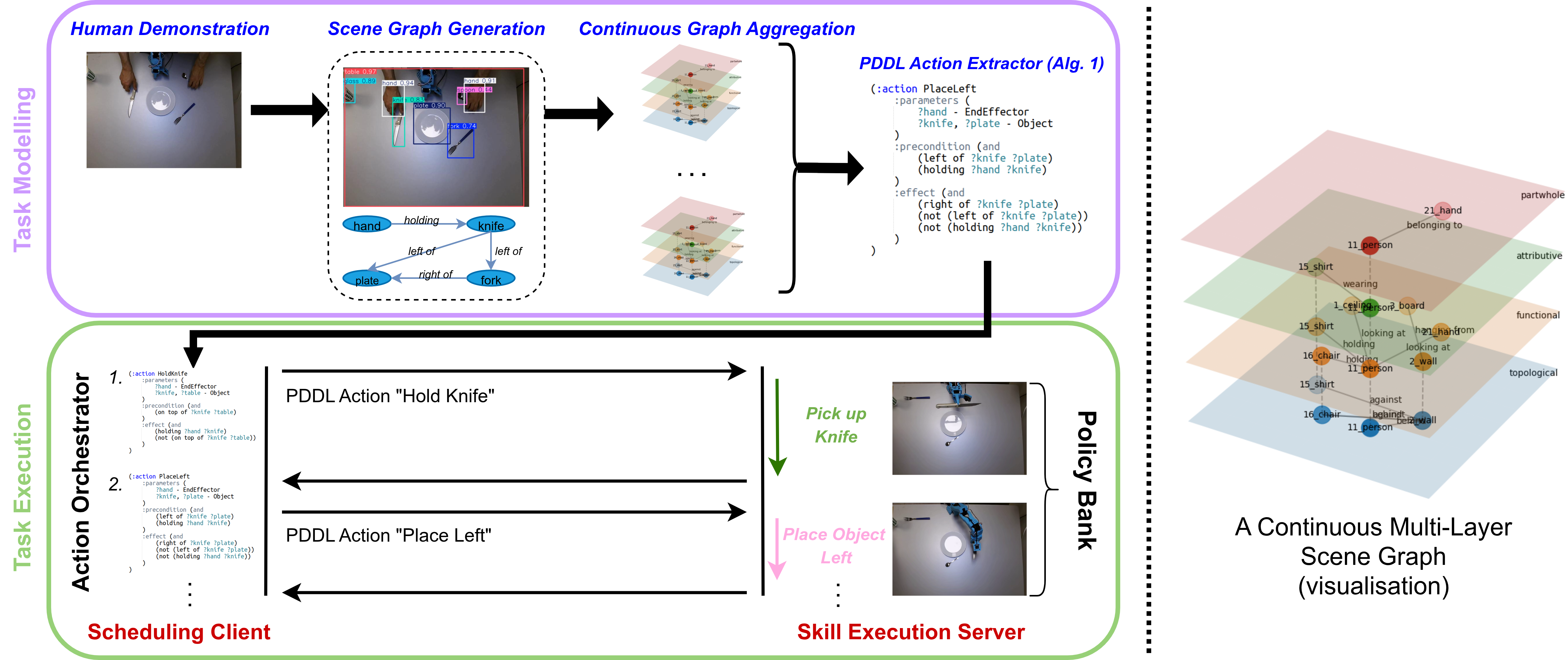}
    \caption{Overall architecture of GraSP-VLA. Top: automatic PDDL action extraction from a single demonstration using Continuous Scene Graph Generation. Bottom: task execution using a bank of pre-trained VLA policies.}
    \label{fig:architecture}
\end{figure*}

\Cref{fig:architecture} provides an overview of the GraSP-VLA architecture.
In Phase I, task modeling takes place (\Cref{fig:architecture} top) by extracting action descriptions from a single demonstration. For that, first the SGG model  is trained on a set of relevant relations to the task (see \Cref{sec:sgg_stage}). Then, during inference, scene graphs are aggregated through time using our Continuous Scene Graph aggregation (see \Cref{sec:continuous_stage}). Finally, the action descriptions in PDDL format are extracted using our PDDL Action Generator (see \Cref{sec:pddl}).

Phase II consists of reproducing the full task, action by action (\Cref{fig:architecture} bottom). For that, the Action Orchestrator (\Cref{fig:architecture} left) asserts preconditions and calls the corresponding VLA policy for each action, using a client-server communication with the Policy Bank (\Cref{fig:architecture} right). The Policy Bank is composed of pre-trained low-level policies described in natural language (for instance, ``Pick up the knife from the table") (see \Cref{sec:vla_policies}).

\subsection{Multi-Layer Scene Graph Generation}\label{sec:sgg_stage}

SGG models represent information from detected entities through bounding boxes and class labels, as well as relations in the form of $\langle subject, predicate, object \rangle$ triplets where $subject$ and $object$ are related to corresponding entity instances.
The traditional SGG paradigm treats various types of triplets similarly. However, visual relations can be of multiple types, which are non-exclusive \cite{zellersNeuralMotifsScene2018a}. 

In previous work of the same authors [anonymized], four different types of relations of interest were defined:
\begin{itemize}
    \item \textit{\textbf{Functional}}: human action or object affordance;
    \item \textit{\textbf{Topological}}: composite (topological + directional) spatial relations between entities;
    \item \textit{\textbf{Physical part-whole}}: hierarchical and invariant relation between a defined entity (i.e. ``whole") and one of its building blocks (i.e. ``part");
    \item \textit{\textbf{Attributive}}: relation between a physical entity and a non-invariant attribute. 
\end{itemize}
Functional relations (e.g. $\langle hand, holding, cup \rangle $) are used to detect related human actions. Topological relations (e.g. $\langle cup, on \ top \ of, table \rangle $) are used to detect object relative spatial states. Physical Part-Whole relations (e.g. $\langle hand, part \ of, person \rangle $) relate specific parts of objects to related actions or states (for instance, the $\langle holding, cup \rangle $ action can be realized through the end effector $hand$). Attributive relations (e.g. $\langle person, wearing, shirt \rangle $) are used to complement the states of objects (for instance, a $shirt$ instance cannot be folded if someone wears it).
Inspired by this categorization, we propose to extend the SGG method to a multi-layer representation. SGG models traditionally give probabilities for every $m$ predicate for every $n \times (n-1)$ number of object pairs, then apply the \textit{argmax} function to extract the top prediction. We improve this by applying the \textit{argmax} function on the subset of predicates from each relation category, leading to a maximum of $n \times (n-1) \times 4$ relations predicted.
It is important to note here that a predicate is not constrained to a specific layer; for instance, the predicate $on$ can be used for a topological relation (e.g.  $\langle cup, on, table \rangle$) and for a functional relation (e.g.  $\langle person, on, computer \rangle$). 
The relation classification is applied to the entire $\langle subject, predicate, object \rangle$ triplet by using a fine-tuned Large Language Model, as in the original work by authors [anonymized]. The resulting multi-layer scene graph (\Cref{fig:architecture}, right) is composed by edges that are layer-dependent and nodes that are shared across layers:
\begin{equation}
    G' = \{V, E, L\}
\end{equation}
where $V$ is the set of vertices, $E$ is the set of edges, and $L$ is the set of layers.
Each node $v \in V$ is represented by $v = (b, c, w)$ where $b$ is the bounding box coordinates, $c$ is the class label, and $w$ is the confidence value. Each edge $e \in E$ is represented as $e = (u, v, l, c, w)$ where $(u,v) \in V$ and $l \in L$. Next, we describe how we extended this representation to the time domain.



\subsection{Continuous Scene Graph}\label{sec:continuous_stage}
On top of the Multi-Layer Scene Graph ($G^{\prime}$), we define a Continuous Scene Graph ($G^+$) structure that is constantly updated with object/relation detections and serves as the internal memory of the robotic agent. 

\begin{definition}[Continuous Scene Graph]
given a set of vertices $V$ and edges $E$ at every discrete time $k$, then:
\begin{equation}
    G^+_k = \{V_k,E_k,f_k,l_k\}
\end{equation}
where $E_k \subseteq V_k^2$; $f_k$ is a function $f_k:E_k \rightarrow F$ that maps edges to their labels and $l_k$ is a function $l_k:E_k \rightarrow L^4$ that maps edges to corresponding layers (i.e. functional, topological, attributive and part-whole).
\end{definition}

\begin{definition}[Updates]
given a continuous SG ($G^+_k$), the function $h$ updates its states  using information collected through the SGG backbone $\gamma$:
\begin{equation}
   h: G^+_k \times \gamma \rightarrow G^+_{k+\gamma}
\end{equation}

\end{definition}

\begin{definition}[Relations]
given a Continuous Scene Graph at a discrete time $k$ for a discrete layer $l$, then the set of relations is defined as:
\begin{equation}   
   \Pi_{(l,k)} = \{ \langle \eta, \mu, f((\eta, \mu)) \rangle \mid \eta, \mu \in V_k, (\eta, \mu) \in E_{lk} \}
\end{equation}
where  a Multi-Object Tracking (MOT) algorithm \cite{caoObservationCentricSORTRethinking2023a} is used to associate a tracking ID $i$ to each object, leading to $v = (b ,c, w, i)$, so that nodes are persistent through time.
To make edges continuous through time, a node pair $(u,v) \in V_k$ is associated a matrix of size $n \times m$, where $n$ is the number of timestamps and $m$ is the number of layers ($m=4$ in \Cref{fig:architecture}). Each cell of the matrix represents a \textit{state} of the relation between two nodes at a given timestamp and for a given layer. 
\end{definition}

\begin{figure}[t]
    \centering
    \includegraphics[width=.99\columnwidth]{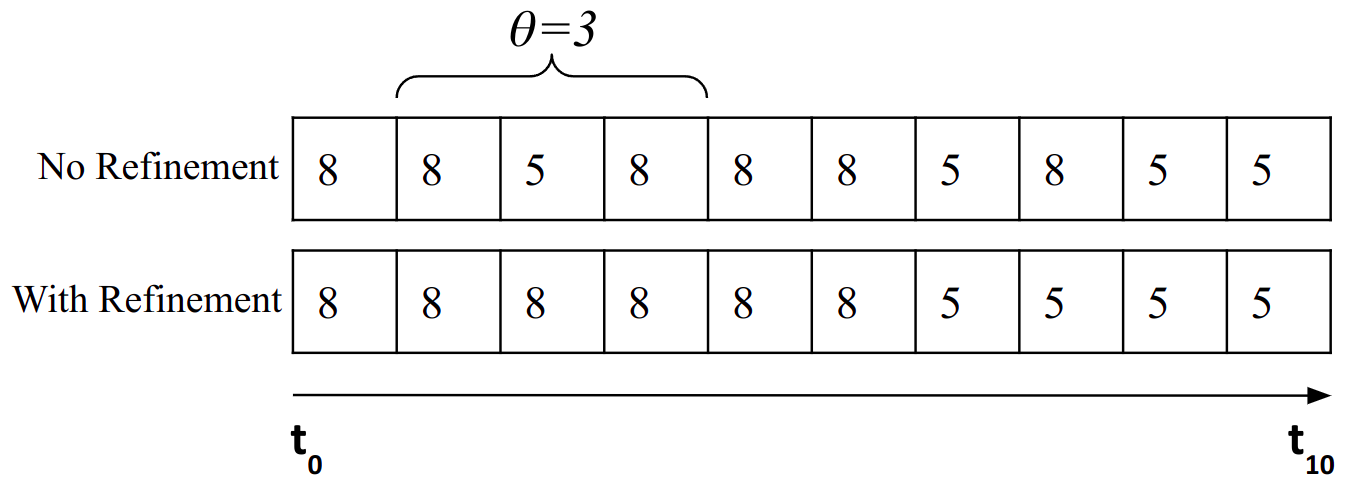}
    \caption[State refinement example]{Example of state refinement for a relation between two nodes at a given layer. States are represented by the label of the relation, for instance, $8 = above$ and $5 = below$. The sliding window is set to 3 timestamps (i.e. $\theta = 3$).}
    \label{fig:state_refinement}
\end{figure}

To make the final graph ($G^+_{k+\gamma}$) more robust, incorrect predictions are also filtered out. So, inspired by previous work \cite{zhuoExplainableVideoAction2019b}, we deployed a state refinement mechanism for this purpose. State refinement works as follows: we set a sliding window variable $\theta$ that represents the number of timestamps to consider for the state refinement. For every new relation detected, we compare it to its previous states and wait for future detections to confirm or refute the relation. We exemplify this process in \Cref{fig:state_refinement}. In the top example (i.e. No Refinement), a false positive for the relation $below$ (id=5) is detected at timestamp 3. If not corrected, this will lead to inconsistency in the internal state of the object, as it is unlikely that an object can be $above$, then $below$, then $above$ another object in a fraction of a second. By using state refinement (i.e. With Refinement in \Cref{fig:state_refinement}), the change of states is validated only if the same label is detected multiple times in a row, leading to a representation that is aligned with real-world constraints. The sliding window is chosen empirically ($\theta = 3$).

In addition, every new relation added to the Continuous Scene Graph ($G^+_{k+\gamma}$) is given a weight value $\omega_r$, which is a value of confidence (or certainty) that the relation exists. When an existing relation is detected again, we update its confidence value as follows:
\begin{equation}
    \omega_r = \omega_{(r-1)} +  \sigma (\tau_{c} - \tau_{r}) ,
\end{equation}
where  $\omega_{(r-1)}$ is the previous confidence value of the relation; $\sigma$ is a constant value ($\sigma = 0.5$); $\tau_{c}$ is the current timestamp; and, $\tau_{r}$ is the last timestamp of the relation. We use the original confidence value given by the SGG backbone ($\gamma$) as the initial weight of the relation. Relations with a low confidence value are automatically removed from the representation.

\subsection{Automatic Action Description Generation}
\label{sec:pddl}

We propose to use our Continuous Scene Graph representation ($G^+_{k+\gamma}$) to autonomously generate a planning domain, which could be further used by the robot to reproduce the observed actions. As a formalism, we use the Planning Domain Definition Language (PDDL) \cite{ghallabPDDLPlanningDomain1998} for describing the domain.
A PDDL domain description is defined by a set of types, predicates, actions, and constants (optional). Predicates in PDDL are defined as relations between two entities in our graph, e.g. $\langle subject, predicate, object \rangle$.
Actions in PDDL are specified by: \textit{Parameters}, the entities involved in the action; \textit{Preconditions}, the conditions that must be true for the action to be executed, and \textit{Effects}, the changes in the environment after the action is executed. In AML, agents are the only source of actions. However, in SGG, there is no distinction in the types of nodes. To solve this issue, we define two types of nodes in our Continuous Scene Graph  ($G^+_{k+\gamma}$) representation: \textit{agent} and \textit{object}. The nodes in the $G^+_{k+\gamma}$ with the label \textit{person} will be defined as \textit{agent} and all other nodes as \textit{objects}. This distinction allows to generate PDDL actions by observing the influence of relations created by agents on other relations.

\begin{definition}[Action]
 an action is defined as the difference between pairs of subsequent sets of relations $\Pi_{k}^-$ and $\Pi_{k+\gamma}^+$, which represent changes $before$ and $after$ an update of the graph, respectively.
\end{definition}
 
\begin{figure}
    \centering
    \includegraphics[width=1.0\columnwidth]{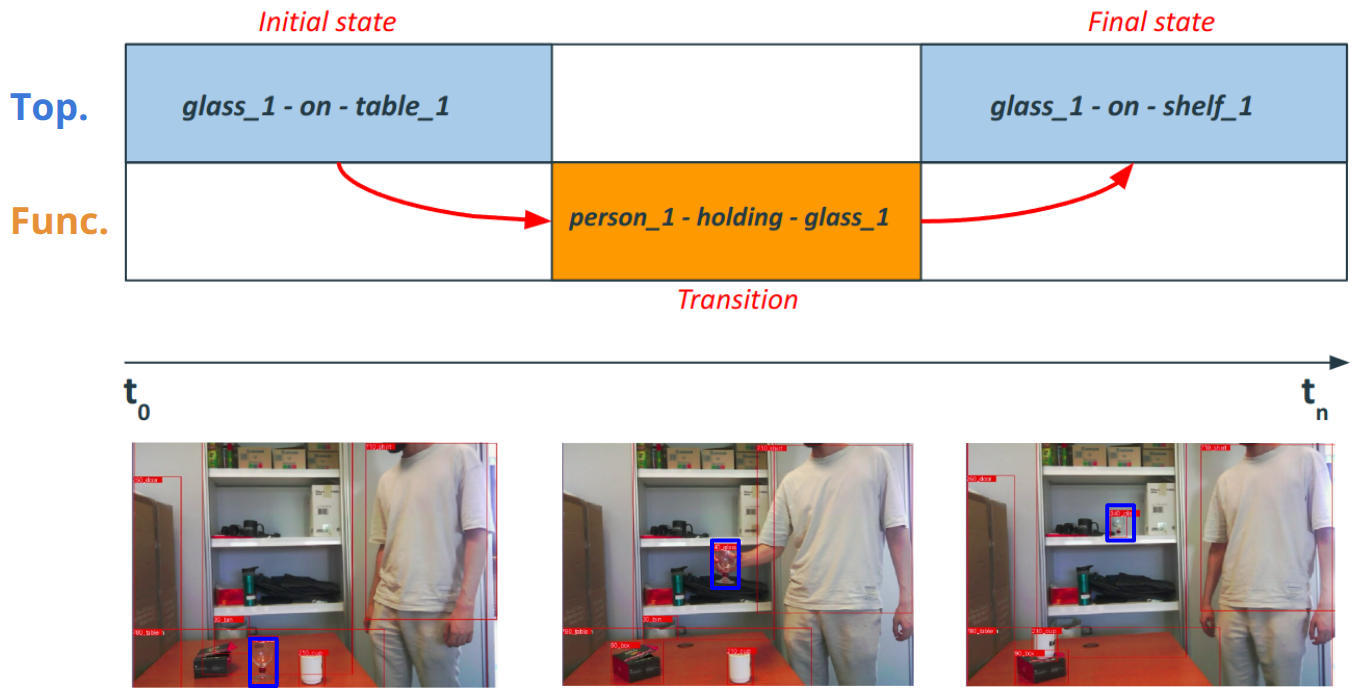}
    \caption[Example of a state transition (diagram)]{Example of a transition identified using the interactions of the \textit{Topological} and \textit{Functional} layers of the Continuous Scene Graph for the action \textit{Moving glass to shelf}.}
    \label{fig:transition_example}
\end{figure}

To illustrate our method for Automatic Action Description Generation, a running example is presented as a simple scenario where a person is grasping a $glass$ from a $table$ and moves it to a nearby $shelf$. \Cref{fig:transition_example} presents a sequence of images of this action, as well as the relevant subset of the Continuous Scene Graph ($G^+_{k+\gamma}$).
On the left, note the initial state of the topological layer, with the relation $\langle  glass\_1, on, table\_1 \rangle $. In the next frame, a person is holding a glass and moving it to a shelf. Note that the relation $\langle person\_1, holding, glass\_1 \rangle$ in the functional layer and the absence of $\langle glass\_1, on, table\_1 \rangle$. Finally, the glass is placed on the shelf (see \Cref{fig:transition_example}, right). Notice that the relation $\langle glass\_1, on, shelf\_1 \rangle $ is identified and $\langle person\_1, holding, glass\_1 \rangle $ disappears. 
To identify preconditions and effects, \textit{transitional states} are defined as actions carried out by an \textit{agent} which modify the state of (at least) one \textit{object}  (e.g. in \Cref{fig:transition_example}, the glass). This is included in the Continuous Scene Graph ($G^+_{k+\gamma}$) as a \textit{Functional} relation which directly leads to a change of one or more \textit{Topological} relations of the same object in the next timestamps. 
Furthermore, note that the preconditions and effects of this action are identified by comparing the states of the Continuous Scene Graph $\Pi_{0}$ and $\Pi_{k}$.
Instead of looking at all the previous states, we set an arbitrary sliding window ($\zeta$) of 10 timestamps to look back and forward for relation changes. Notice that selecting an inappropriate number of states, determined by $\zeta$, can either result in excessive memory consumption or yield sub-optimal outcomes.

In \Cref{alg:planning}, we describe our proposed algorithm to find preconditions and effects for a given transition in the Continuous Scene Graph ($G^+_{k+\gamma}$) and generate the corresponding PDDL action. This algorithm is applied at every new timestamp on all \textit{Functional} relations detected at time $k=n$. Then, preconditions $\text{P}$ are identified in corresponding \textit{Topological} relations from time $k=n-\zeta$ to $k=n$ and effects $\text{E}$ in corresponding relations from time $k= n$ to $k =n + \zeta$. If at least one relation is different between preconditions and effects, then a PDDL action is generated and kept in memory.
In addition, the relative complement of the preconditions over effects $\text{NE}$ is included as a set of negative effects of the action (e.g. {\fontfamily{qcr}\selectfont not(holding(hand,cup))} is added to effects in the previous example in \Cref{fig:transition_example}).

\begin{algorithm}[t]
\caption[Preconditions and effects identification]{Automatic preconditions and effects extraction.} \label{alg:planning}
\begin{algorithmic}[1]
\Require Sliding Window $\zeta$, Functional layer $\mathcal{F}$, Topological layer $\mathcal{T}$
\Ensure $\text{List of Actions} \ \mathcal{A}$

\State $\mathcal{A} \gets \emptyset$

\ForAll{$r_f=\langle s,p,o \rangle \in \Pi_{(\mathcal{F},k)}$}
    \State $\text{P}(r_f) \gets \{ \langle s_1,p_1,o_1 \rangle \in \Pi^-_{(\mathcal{T},k-\zeta)} \mid s_1=o \}$
    \State $\text{E}(r_f) \gets \{ \langle s_2,p_2,o_2 \rangle \in \Pi^+_{(\mathcal{T},k+\zeta)} \mid s_2=o \ \wedge\ \langle s_2,p_2,o_2 \rangle \notin \text{P}(r_f) \}$

    \If{$\text{P}(r_f)\neq\emptyset \ \wedge\ \text{E}(r_f)\neq\emptyset \wedge\ \text{E}(r_f)\neq\text{P}(r_f)$}
    \State $\text{NE}(r_f) \gets \{ not(r) \mid r \in \text{P}(r_f) \ \wedge\ r \notin \text{E}(r_f) \}$
    \State $\text{E}(r_f) \gets \text{E}(r_f) \cup \text{NE}(r_f)$
        \State $\mathcal{A} \gets \mathcal{A} \cup \{\langle r_f,\text{P}(r_f),\text{E}(r_f)\rangle\}$
    \EndIf
\EndFor

\State \Return $\mathcal{A}$
\end{algorithmic}
\end{algorithm}

\subsection{Scheduling of VLA policies}\label{sec:vla_policies}

During inference, we extract PDDL action descriptions using the Continuous Scene Graph ($G^+_{k+\gamma}$). In addition to the action descriptions, the order (scheduling) of the actions is arranged by the \textbf{\textit{Action Orchestrator}}. Each PDDL action is mapped (if possible) to known skills using the predicate and object involved, then called through the skill execution server to reproduce the entire task, skill by skill.
It is important to note here that the duration of each skill is not preserved during execution. In the \textit{Open-Ended Imitation Learning} paradigm, we do not have access to a defined goal; hence our approach does not use a dedicated solver. The preconditions of each action are validated using the current state of the $G^+_{k+\gamma}$ before execution of the given policy. If the preconditions are not satisfied, the next policy is called.

When a policy is called, observations are sent to the \textbf{\textit{Policy Bank}} through client-server communication. The communication is synchronous, meaning that the current execution is stopped if a new policy is called. To maintain real-time latency, all available policies are loaded in memory in advance. When the execution of a policy is over, a signal is sent back to the client, and the next policy is called if the corresponding action has been decoded.


\section{Experiments}

Three different sets of experiments were carried out to evaluate our approach, both on datasets and in real-world settings. This section first describes the training and evaluation of the SGG backbone (\Cref{sec:sg_expes}). Secondly, it presents the validation of the  proposed Continuous Scene Graph aggregation and our Action Description Generation algorithm on
the DAily Home LIfe Activity (DAHLIA\cite{vaquetteDAilyHomeLIfe2017}) dataset
(\Cref{sec:pddl_expes}). Third, it describes the evaluation of the entire GraSP-VLA architecture in a final set of real-world experiments using the SO-101 robot arm (\Cref{sec:real_world_expes}).

\subsection{Scene Graph Generation Evaluation}\label{sec:sg_expes}

For SGG, we used the REACT model [anonymized] with the YOLOV8m backbone \cite{Jocher_Ultralytics_YOLO_2023} for object detection.
We trained the REACT model on the IndoorVG dataset [anonymized], which is the only publicly available SGG dataset designed to represent human actions in indoor settings. IndoorVG is composed of 84 object classes and 34 predicate classes.
We trained the YOLOV8 object detector first, and then the REACT model with the object detection part frozen, following original hyperparameters and training details [anonymized]. 
Following previous work \cite{tangLearningComposeDynamic2019}, we used the Recall@K (R@K) and meanRecall@K (mR@K) metrics to measure the performance of models. R@K and mR@K metrics evaluate the top $K$ ($k=[20,50,100]$) relations predicted, ranked by confidence. 
R@K evaluates the overall performance of a model on the selected dataset, whereas mR@K evaluates the performance on the average of all predicate classes, which is more significant for long-tail learning, such as in the task of SGG. 
Latency is measured with batch size 1\protect\footnotemark.
\footnotetext{Hardware: \text{11th Gen Intel\textsuperscript{TM}} Core\textsuperscript{TM} i9-11950H @ 2.60GHz x 16, NVIDIA GeForce RTX 3080 GPU Laptop 16GB, 32GB 3200 MHz RAM.}

\begin{table}
    \centering
    \caption[Real-world performance of the REACT model]{Performance of the REACT SGG model on the IndoorVG dataset, with $\alpha = 0.194$.}
    \begin{tabular}{c|c|c|c|c}
         \multirow{2}{*}{\textbf{Model}} & \textbf{R} & \textbf{mR} & \textbf{mAP\textsuperscript{50}} & \textbf{Latency}\\
         & \textbf{@50/100} & \textbf{@50/100} &  & \textbf{(ms)}\\
        \hline
        REACT & 31.4 / 34.5 & 17.5 / 19.7 & 37.9 & 26.6 \\
    \end{tabular}
    \label{tab:sgg_performance}
\end{table}

 \Cref{tab:sgg_performance} shows our results. In order to lower the number of false positives in object detection, we applied a threshold $\alpha$ to filter out low-confidence detection. We observed that relation prediction metrics (R@K and mR@K) are relatively low which can be explained by the overall complexity of the task, note that  state-of-the-art models in SGG hardly perform better than 20 mR@100 on leading datasets \cite{imEGTRExtractingGraph2024}. More concerning, we observed struggles for the model to attain good accuracy on fine-grained action-related predicates (for instance, Recall@100 for $holding$ is 0.29). This issue is well-known in the SGG community \cite{tangUnbiasedSceneGraph2020}, so in the next section we further explore  how this issue might impact the performance of the subsequent parts in our framework.

\subsection{Continuous Scene Graph \& Action Descriptions}\label{sec:pddl_expes}

\begin{table}
    \centering
    \caption{Results for planning domain generation on the DAHLIA dataset. TP = True Positives, FP = False Positives.}
    \begin{tabular}{c|m{3em}|m{3em}|m{3em}|m{3em}}
        \multirow{2}{*}{\textbf{Video}} & \multicolumn{2}{c|}{\textbf{Baseline}} & \multicolumn{2}{c}{\textbf{w/ Informative}} \\
         & \textbf{TP} & \textbf{FP}  & \textbf{TP} & \textbf{FP} \\
        \hline
        \textbf{1} & 3 & 0 & 21 & 14  \\
        \textbf{2} & 1 & 0 & 19 & 28 \\
        \textbf{3} & 3 & 1 & 59 & 44  \\
        \textbf{4} & 1 & 2 & 19 & 18  \\
        \textbf{5} & 1 & 1 & 43 & 51  \\
        \hline
        \hline
        \textbf{Recall} & \multicolumn{2}{c|}{0.69} & \multicolumn{2}{c}{0.51}
    \end{tabular}
    \label{tab:dahlia_planning}
\end{table}

To track objects and make nodes in the graph continuous, we used the OC-SORT Multi-Object Tracker \cite{caoObservationCentricSORTRethinking2023a} on top of the bounding box predictions of REACT.
To measure the performance of our automatic Action Description Generation approach, we used the DAily Home LIfe Activity (DAHLIA) dataset \cite{vaquetteDAilyHomeLIfe2017}. DAHLIA is a dataset of long-term human activities performed in home environments, which is composed of 44 videos of 44 different subjects performing daily life activities. This dataset contains 7 different annotated activities, such as \textit{cooking} or \textit{washing dishes}. 
We evaluated the relevance of our Automatic Action Description Generation (see next \Cref{sec:pddl}) by measuring the number of correct actions identified and translated into PDDL. We focused on actions including the predicates \textit{holding} and \textit{using}. Our approach was evaluated on 5 different videos randomly sampled in the DAHLIA dataset, with an average of 41' each.
Since DAHLIA does not contain any ground truth for the scene graphs or actions, we had to manually evaluate the relevance of the generated PDDL actions as follows: (i) for each PDDL action generated, we watched the corresponding video clip and evaluated if the action was correctly identified and translated into PDDL, and then (ii) we computed the Recall of the approach as the number of correct actions over the total number of actions generated.

Out results are presented in \Cref{tab:dahlia_planning} (Baseline). We observed a high Recall but a very low number of actions identified, with only an average of 2.6 actions per video. As mentioned in \Cref{sec:sg_expes}, the SGG model struggles to identify the action-related fine-grained predicates (such as $holding$), preferring coarse-grained ones such as $next \ to$ or $on$. To overcome this issue, we propose to use the \textit{Informative Selection} approach for SGG [anonymized]. The Informative Selection re-weights predictions of an SGG model according to semantic importance. With this method, relations predicted are more informative but with lower confidence. \Cref{tab:dahlia_planning} (w/ Informative) shows the results after applying \textit{Informative Selection}. Note that a consequently higher number of actions are generated, mainly because more action-related predicates are predicted. The high number of False Positives in this setting can be attributed to (1) wrong or missing object detection and (2) low recall for relation prediction (as explained in \Cref{sec:sg_expes}).

\subsection{Evaluation of GraSP-VLA}\label{sec:real_world_expes}

\begin{figure}
    \centering
    \begin{subfigure}{0.32\columnwidth}
        \centering
        \includegraphics[width=\columnwidth]{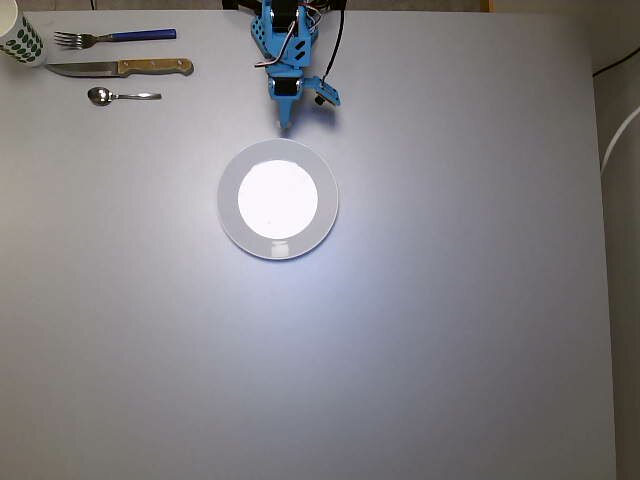}
        \caption{}
        \label{fig:initial_pose}
    \end{subfigure}
    \begin{subfigure}{0.32\columnwidth}
        \centering
        \includegraphics[width=\columnwidth]{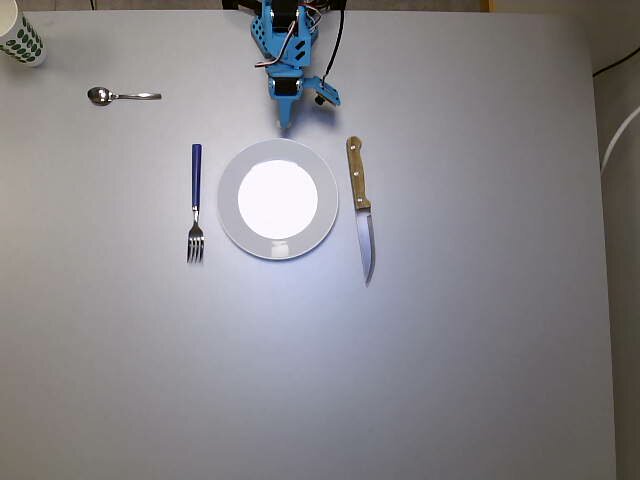}
        \caption{}
        \label{fig:end_pose1}
    \end{subfigure}
    \begin{subfigure}{0.32\columnwidth}
        \centering
        \includegraphics[width=\columnwidth]{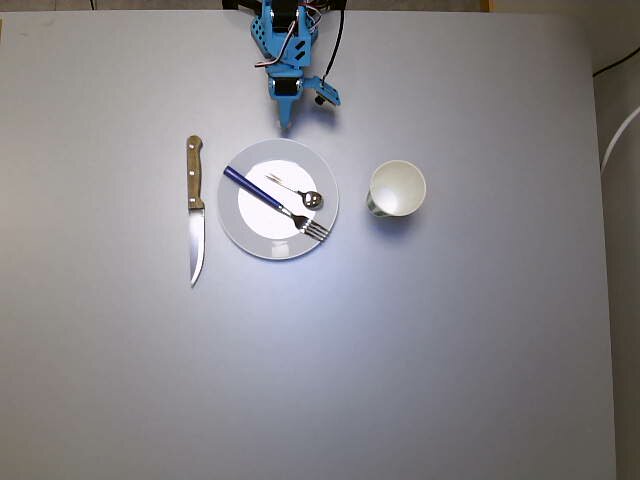}
        \caption{}
        \label{fig:end_pose2}
    \end{subfigure}
    \caption[Example of a state transition (figure)]{(a): initial setup; (b) - (c): possible end configurations.}
    \label{fig:setting_table}
\end{figure}

In a final set of experiments, we evaluated the entire GraSP-VLA, from human demonstrations to action description generation to real-world execution with VLA policies. 
We created a simple scenario that consists of setting up a dinner table. 
In the middle of the operation area, we placed a plate (static). Then, a fork, knife, and spoon are disposed in a loading area, waiting to be placed around the plate by the demonstrator (see \Cref{fig:initial_pose}). The human demonstrator can pick and place any object in any order, with three different spatial goals: \textit{left of} the plate, \textit{right of} the plate, or \textit{inside} the plate. 
In this scenario, the complexity of the skills themselves is low (e.g. grasping and placing simple objects); however, the number of skill combinations can grow nearly infinite, as objects can be replaced by each other, increasing the complexity of the task as the number of pick-and-places grows. Our goal with these experiments is to show that our approach can still maintain good accuracy even when the length of the task increases.
We evaluated different complexities, from 1 pick-and-place in a row to 6 pick-and-places in a row (each pick-and-place can contain the same or different objects). Examples of end configurations are displayed in \Cref{fig:end_pose1} and \Cref{fig:end_pose2}.
We used the SO-101 robot arm \footnote{\url{https://github.com/TheRobotStudio/SO-ARM100}} for our experimentation and we collected a dataset of 20 demonstrations, where each demonstration consists of moving each object once to each possible end-location. Using these data, we fine-tuned the SmolVLA model \cite{shukor2025smolvla} for 6 policies, as follows: \textit{pick\_knife, pick\_fork, pick\_spoon, place\_left, place\_right, place\_inside}. To train these policies, each demonstration segment was cut according to the current stage of the demonstration. Each policy has been trained for 20,000 steps with a batch size of 64 using the original code and the same hyperparameters used by Shukor et al. \cite{shukor2025smolvla}.

This simpler scenario was designed to measure the success rate of our proposed  \Cref{alg:planning} when extracting PDDL actions in real-time. To detect spatial relations in the scene, we applied Qualitative Spatial Reasoning (QSR) and used a reference system (similar to that defined by Moratz and Ragni \cite{moratz2008qualitative}) centered on the plate. This resulted in 6 different relation predicates detected between objects: \textit{right of, left of, above, below, inside}, and \textit{holding}. Here, we kept the object detector of the REACT model but removed the relation prediction head to extract only bounding-box coordinates for each object. Then, relations are populated using object coordinates with respect to the reference frame. To detect the relation $\langle hand, holding, object \rangle$ we used the Intersection over Union (IoU) approach applied on  the hand bounding-box and objects, removing duplicates if necessary.

\Cref{tab:acc_smolvla} shows the obtained accuracy after executing each independent policy 10 times, starting from the initial configuration with a maximum execution time of 60''. Note that the \textit{Placing} policies are correctly learned, whereas \textit{Picking} policies have a lower success rate. This can be due to the low number of recorded episodes (20) when grasping each object. Since the \textit{Placing} policies are shared by each object, the number of training episodes is higher for spatial locations (3 objects x 20 demonstrations = 60 episodes).
\begin{table}[]
    \centering
    \caption{Accuracy of the SmolVLA \cite{shukor2025smolvla} policy for each atomic behavior.}
    \begin{tabular}{c||ccc|ccc}
        \hline
        \multirow{2}{*}{\textbf{Policy}} & \multicolumn{3}{c|}{\textbf{Pick}} & \multicolumn{3}{c}{\textbf{Place}} \\
          & \textit{knife} & \textit{fork} & \textit{spoon} & \textit{left} & \textit{right} & \textit{inside} \\
         \hline
         \textbf{Acc.} & 0.8 & 0.6 & 0.7 & 1.0 & 1.0 & 1.0 \\
         \hline
    \end{tabular}
    \label{tab:acc_smolvla}
\end{table}

\Cref{tab:task_results} reports the accuracy of our entire approach (including Orchestrator and Skills Execution) on a set of short- to long-horizon tasks. For each task, we chose a random end-configuration and scheduling. Each end-configuration was strictly different from the ones used during the training of the VLA policies. Note that \textit{skill} refers to the combination of one pick-and-place action on a single object. First, we report the accuracy of generating correct action-representations in PDDL (\textit{Action Acc.} in \Cref{tab:task_results}). 
Note a net improvement compared to the previous experiments, with an average of 0.96\% of accuracy to generate the correct PDDL action descriptions. Regarding the overall accuracy of the tasks (\textit{Acc.} in \Cref{tab:task_results}), we observe a decrease in the number of skills chained in a row. This is due to the low accuracy of certain skills (for instance \textit{pick(fork)} in \Cref{tab:acc_smolvla}).
\begin{table}[]
    \centering
    \caption{Success rate of our GraSP-VLA architecture during real-world experiments.}
    \begin{tabular}{c||c|c|c}
        \hline
         \multirow{2}{*}{\textbf{\# Skills}} & \multicolumn{2}{c|}{\textbf{GraSP-VLA}} & \textbf{Full Fine-Tuning} \\
         & \textbf{Action Acc.} & \textbf{Acc.} & \textbf{Acc.} \\
         \hline
         \textbf{2} & 1.0 & 0.6 & 0.2 \\
         \textbf{4} & 1.0 & 0.4 & 0.1 \\
         \textbf{6} & 0.9 & 0.4 & 0.0 \\
         \hline
    \end{tabular}
    \label{tab:task_results}
\end{table}

We also compared our approach to a full fine-tuning of SmolVLA on the entire task, with a combination of 2 to 6 skills, again with an average of 10 test episodes. For this fine-tuning, we performed each task with a unique scheduling for 20 demonstrations. Last column in \Cref{tab:task_results} shows our results (\textit{Full Fine-Tuning}). We observe a very low success rate, highlighting the limitations of the SmolVLA model with long tasks and a low number of demonstrations. By decoupling the learning in sets of unitary behavior and using our Continuous Scene Graph ($G^+_{k+\gamma}$) to decompose the task into a sequence of actions, we are able to largely improve the success rate. For instance, our approach improves the accuracy from 0.2 to 0.6 when combining 2 skills in a row. It is important to note that our approach was tested in more challenging settings than the baseline \cite{shukor2025smolvla}, since a different scheduling was used for each demonstration.

\section{Discussion}

The main limitation of our GraSP-VLA approach lies in the SGG model. As discussed in \Cref{sec:sg_expes} and \Cref{sec:pddl_expes}, SGG models struggle to predict relations with high accuracy, leading to a consequent amount of False Positive actions. We believe that, with a larger and more diverse baseline dataset, GraSP-VLA may identify more actions in the scene. Moreover, the transfer learning from the IndoorVG to the DAHLIA dataset can explain the limits of the SGG model, since the quality and diversity of images are different.

The main strength of  GraSP-VLA is that it can be extended in many ways. For instance, in addition to action descriptions, it can infer new relations between object classes by aggregating actions generated by \Cref{alg:planning}.
To follow up on our experiments in \Cref{sec:pddl_expes}, we can learn a new class of entities called \textit{movable} by aggregating all actions that involve moving an object from one place to another. The action of ``moving'' is determined by a change of relation in the \textit{Topological} layer of the Continuous Scene Graph ($G^+_{k+\gamma}$), as seen in \Cref{fig:transition_example}. By identifying objects involved in the effects of such PDDL actions, we can define a new object feature, ``movable''; and, objects involved in the preconditions but never in the effects of an action can be featured as ``static'' objects. By mining these two types of objects from the DAHLIA dataset, we obtained a realistic list (see \Cref{tab:classes}). This showcases the possibilities of GraSP-VLA for extracting not only action descriptions but also ontologies from our Continuous Scene Graph ($G^+_{k+\gamma}$) representation.

Finally, compared to traditional VLA training, our GraSP-VLA architecture is more flexible, as it allows the learning of a new task from a single demonstration if all skills used in the task are present in the policy bank. The second advantage is that the outcome of each policy execution can be monitored using the Continuous Scene Graph ($G^+_{k+\gamma}$). If all relations present in the effect of the PDDL action are not detected after the completion of the policy, the skill can be called again.
This type of fallback mechanism can significantly improve the overall success rate of complex tasks.

\begin{table}
    \centering
    \caption[Ablation study: ontology building]{Classes of entities extracted from the DAHLIA dataset.}
    \begin{tabularx}{0.8\linewidth}{X|X}
        \textbf{Movable Objects} & \textbf{Static Objects} \\
        \hline
        bottle, door, bag, basket, bowl, cup, knife, glass, plate & cabinet, microwave, counter, sink, faucet, floor, wall, table, shelf
    \end{tabularx}
    \label{tab:classes}
\end{table}


\section{Conclusion}

This paper introduced a new approach, GraSP-VLA, for symbolic planning with a Continuous Scene Graph in Imitation Learning. This approach models a new representation of the environment called the Continuous Scene Graph ($G^+_{k+\gamma}$), which represents the evolution of compositional relations over time. This representation is powered by a state-of-the-art SGG backbone and a Multi-Object Tracking algorithm. In contrast to standard representations of this sort, we propose to divide our representation into four different layers, each representing a category of relations: \textit{Topological}, \textit{Functional}, \textit{Part-Whole}, and \textit{Attributive}. We used state refinement to filter out wrong predictions and improve the stability of the graph through time.
In addition, we proposed a new algorithm to generate planning domains from this representation. Planning domains are important tools for learning the symbolic representation of new skills by autonomous robots. We showed through experiments on the DAHLIA dataset that our approach can extract action descriptions from real-world videos without the need for explicit priors.
Finally, we proposed using our Continuous Scene Graph for scheduling low-level VLA policies in single-shot demonstration learning. We deployed our approach in a set of real-world experiments, using a client-server architecture to schedule the execution of policies. Results show the superiority of our approach compared to classical training on the full task. 

Our GraSP-VLA architecture decomposes the symbolic representation of the task and the execution of VLA policies as separate components. However, recent approaches infuse external signals such as object representations during fine-tuning of VLA policies \cite{li2025controlvla}. As future work, we intend to extend our approach by infusing a latent representation of the Continuous Graph as an additional signal for the Action Expert of the VLA model. We believe that the spatial information contained in key graph relations can increase the success rate of VLA policies in real-world settings. 

\addtolength{\textheight}{0pt}   





\section*{ACKNOWLEDGMENT}

Zoe Falomir and Maëlic Neau acknowledge the Knut and Alice Wallenberg foundation and the Wallenberg AI, Autonomous Systems and Software Program (WASP).

\bibliographystyle{IEEEtran}
\bibliography{IEEEabrv,references}


\end{document}